# Impactful Robots: Evaluating Visual and Audio Warnings to Help Users Brace for Impact in Human Robot Interaction

Nathaniel G. Luttmer[1], Takara E. Truong[2] Alicia M. Boynton[3], Andrew S. Merryweather[1], David R. Carrier[3], and Mark A. Minor[1], IEEE Member

*Abstract*— Wearable robotic devices have potential to assist and protect their users. Toward design of a Smart Helmet, this article examines the effectiveness of audio and visual warnings to help participants brace for impacts. A user study examines different warnings and impacts applied to users while running. Perturbation forces scaled to user mass are applied from different directions and user displacement is measured to characterize effectiveness of the warning. This is accomplished using the TreadPort Active Wind Tunnel adapted to deliver forward, rearward, right, or left perturbation forces at precise moments during the locomotor cycle. The article presents an overview of the system and demonstrates the ability to precisely deliver consistent warnings and perturbations during gait. User study results highlight effectiveness of visual and audio warnings to help users brace for impact, resulting in guidelines that will inform future human-robot warning systems.

## I. INTRODUCTION

As humans and robots interact with each other, it is important to understand how to best couple them together to communicate dangerous situations. Wearable robots equipped with advanced sensory systems can supply a warning and prepare humans for an event they are about to experience. Warning systems are used in different applications to give humans feedback about possible events. The automotive industry has used visual, auditory, haptic, and augmented reality to warn humans about possible collisions [1] [2]. Auditory warnings have been used for workers in hazardous environments to warn them of dangerous situations [3].

This work is motivated by the development of a Smart Helmet to protect users from brain injury while playing sports. The helmet will use radar to detect impending collisions [4], trigger warnings to help the user brace for the impact, and then use controllable air bladders inside the helmet to dissipate impact forces [5].

To protect humans, reflex muscle activation cannot occur quickly enough, hence precisely timed warnings are required so that muscles generate forces in time to protect against impact, but not so far in advance that they affect player performance. Auditory warnings have been studied using a four tethered head impulse system, which highlighted the role of cervical muscle activation with a seated user [6].

*Research supported by the National Science Foundation under Grant No. 1622741.

Auditory warnings were also examined in [7] in a small pilot study to evaluate the ability of users to brace for impact while running.

This article builds upon our prior work [7] to provide improved warnings of simulated impacts as users are running. Combinations of audio and visual warnings are examined to determine which results in the best user performance. While this work centered around a Smart Helmet, the results of this study can help other researchers design warning systems for other human-robot interaction (HRI).

*A. Approach*

To test different kinds of warning systems a safe and controlled environment is needed. This article uses the same system as [7] to precisely generate simulated impacts, but augments the tether-based system with new visual warnings and validates that the robot performance is appropriate for a broader subject study.

To quantify the effectiveness of a warning, the participant experiences perturbation forces from randomized directions with and without warning. The warning systems indicates which direction the participant will experience those forces while providing participants enough time to clench core muscles before they are perturbed. The no-warning perturbations give a baseline participant response. For this work, the perturbations occur during the stance phase of a participant's locomotor cycle (i.e., stride) to allow foot to ground contact so they can resist the incoming perturbation.

The TreadPort Active Wind Tunnel (TPAWT) [8] is used in this research. The TreadPort has a large treadmill (1.8 x 3.25 m) with a rigid tether to allow users to move freely while traversing a virtual world and experiencing locomotion forces. The TreadPort is ideal for this study since it provides sufficient space for perturbed user motion and has resources required for conducting these experiments. The graphics are disabled to remove distractions, and the rigid tether is not used since it impedes user motion during running and perturbations. A 3-wire body weight support system from [9] was adapted for the perturbation study [7] to apply horizontal force to the users' hips, Fig. 1.

VICON motion tracking inside the TPAWT provides coordination of the user study. Motion tracking provides warnings and perturbation forces coordinated with the user's stride. Markers on the user provide gait monitoring, which is used to predict the stance phase during a stride. Auditory and visual warnings are thus triggered moments before the perturbation is applied to allow time to prepare prior to a perturbation. Tether forces are calculated in real time to produce the desired perturbation force based upon the position of the participant on the treadmill with respect to the tether standoff positions. Motion tracking measures participant displacement after perturbations to characterize the effectiveness of different warnings

Both audio and visual warnings are studied in this article. Audio warnings are produced by buzzers positioned peripherally around the user, although only the front buzzer is visible in Fig. 1. A visual warning using an LED display is positioned in front of the user where different LEDs are illuminated to indicate the direction of the perturbation. Given the gait monitoring system, warnings and perturbations are synchronized with the participant's gait cycle.

System performance is validated to assure that it provides consistent operation before conducting the user study. For an effective system, tether vibration is examined to find a static tether tension that prevents tether vibration during running while not impeding participant motion or limiting perturbation force. Perturbation forces need to scale with user mass to provide appropriate forces. Hence, this article examines how scaling affects the attainable workspace of the system to assure that forces can be applied at different user positions. Sufficient bandwidth for applying perturbations is assured by modelling the system and analyzing its dynamics when supplying a tether force. The full system logic is presented to illustrate autonomous system operation and control leading to fully randomized subject studies with consistent operation.

It is important to identify effective warnings so that people can prepare for unexpected events. A user study is performed to understand how humans respond to various combinations of warning types while running. This is accomplished by analyzing the displacement of perturbed participants while running. Perturbations are applied during stance so the user can compensate by adjusting ground reaction forces to reduce the disturbance to their gait. To evaluate how warnings affect the response to a perturbation, a user study was designed to study warning type (audio, visual, combined audio-visual, or no warning). Six participants were recruited and Linear Mixed Effect (LME) models were used to investigate the significance of the participant's displacement from the perturbation during each warning type.

B. *Literature Review*

Various warning modalities have been examined to determine human response. Trunk muscle activity has also been studied when perturbed while standing [10]. Anticipation of direction and time of a perturbation [11] has been studied to understand trunk muscle activation while the participant is sitting. Auditory warning sounds have been studied for hazardous work environments, highlighting that a limited number of different tones and melodies help users understand the warning [3]. Tones emitted at spatial locations around the user warned of perturbation directions in [7], where a different tone was used behind the user. Tones were effective for bracing for perturbations, but some users still had difficulty discerning front/back directions despite the different tones. Directional audio warnings and non-directional startling sounds, both using headphones, were examined in [6] with a seated user bracing for head perturbations. Results indicate that directional audio helped the user best brace for head perturbations, followed by startle warnings.

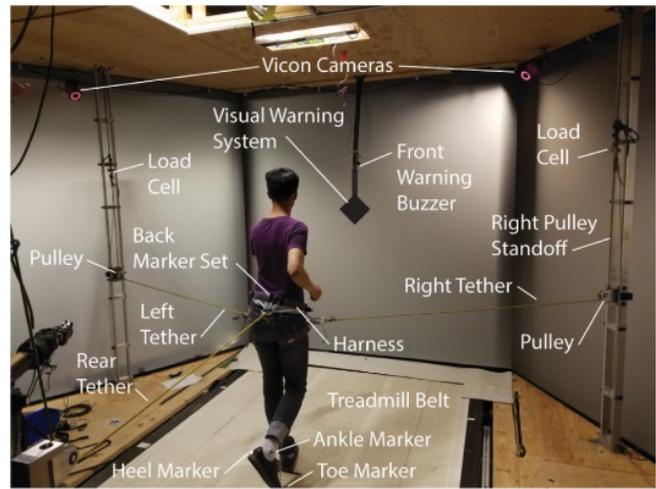

Fig. 1. Perturbation System Overview.

VR was studied as a means of understanding user response to visual and audio warnings, highlighting that more realistic simulations are important [1]. Audio and visual warnings have also been studied in simulators examining intersection collision warning systems, highlighting that drivers can better identify dangerous vehicles and take corrective actions [2]. This article studies different combinations of audio and visual warnings while a user is running to develop a better understanding of how to help users brace for impact.

Other researchers have examined the effect of perturbations while the user is moving. Researchers examined trunk muscle activation during walking with forward and backward perturbations [12]. Studies examined participant balance after an unexpected perturbation using a locomotion interface [13]. Tethered systems with a treadmill have been utilized to apply a perturbation [14], but to the best of our knowledge, related work does not include real-time gait tracking or a predictive warning system to apply perturbations. Their forces are limited to 200 N [14], whereas the system presented here is shown to reliably produce good workspaces with 400 N forces, sufficient for notable perturbations and displacements. Pinpoint application of forces and a priori warnings in this work are possible due to predictive gait tracking. Gait tracking has been researched for pinpointing toe-off and heel-strike [15-18], but this work examines stance prediction, which was originally derived in [7]. Of the related research, only [7] examined perturbation displacements related to warnings and to our knowledge none have examined combinations of audio and visual warnings.

*C. Contributions*

This article demonstrates the effect of audio and visual warning systems to help users brace for impact during running. This information is helpful for designing HRI systems where robots help protect humans during physical activity. While all warnings are proven to be beneficial, visual warnings are proven to be best in terms of overall reduction of displacements and statistical significance of the improved performance. This work also validates the TPAWT modifications made to reliably perform perturbation studies while running. This is possible because of predictive gait tracking, timing of warnings and force application, and sufficient system dynamics. The later assures forces and impulses are well controlled, despite user motion and weight, to result in a large workspace and reliable force application. This information is helpful to other researchers designing systems for evaluating dynamic human robot interaction.

## II. PERTURBATION SYSTEM

*A. Mechanical Design*

A body weight support system designed for the TPAWT is adapted to provide perturbation forces [9]. The perturbation system uses three motor-controlled tethers to apply forces to perturb the user. Drive motors are located in the ceiling, Fig. 2 (a). Each motor can apply up to a 600 N force, sufficient for the desired perturbation. Transducer Technique load cells, Fig. 2 (b), measure the actual tether forces, providing a resolution of 13.58 mN and a 1kHz bandwidth. A dSpace system running at 10 kHz is used to compute all of the necessary inputs and outputs to control the tethers and read sensors. A second dSpace system is used to control belt speed.

Each of the tethers is routed through an articulated pulley system to the participant's body via a waist harness. A specially designed articulated back pulley, which can orient the cable around the rigid tether used in prior work was installed Fig. 2 (b). The standoffs route the tethers after they exit the ceiling, Fig. 2 (c), to make the system a center of gravity horizonal perturbation system. Pulleys on the right and left side of the participant are supported at hip-height by standoffs so they can swivel relative to the position of the user, Fig. 2 (c).

*B. Tether Force Calculations*

The forces from each of the tethers, $F_{LT}, F_{RT}, F_{BT}$, are combined to create the desired perturbation force, $F_{des}$, and perturbation angle, β. Fig. 3 outlines the positions of the origin, $O_o$, the positions of the tether standoffs with respect to the origin, $O_{LT}, O_{RT}, O_{BT}$, the position of the participant, $O_B$, and the position of the participant after perturbation, $O_P$. Using the positions of the standoffs and the participant, the force for each of the tethers are calculated to give the desired

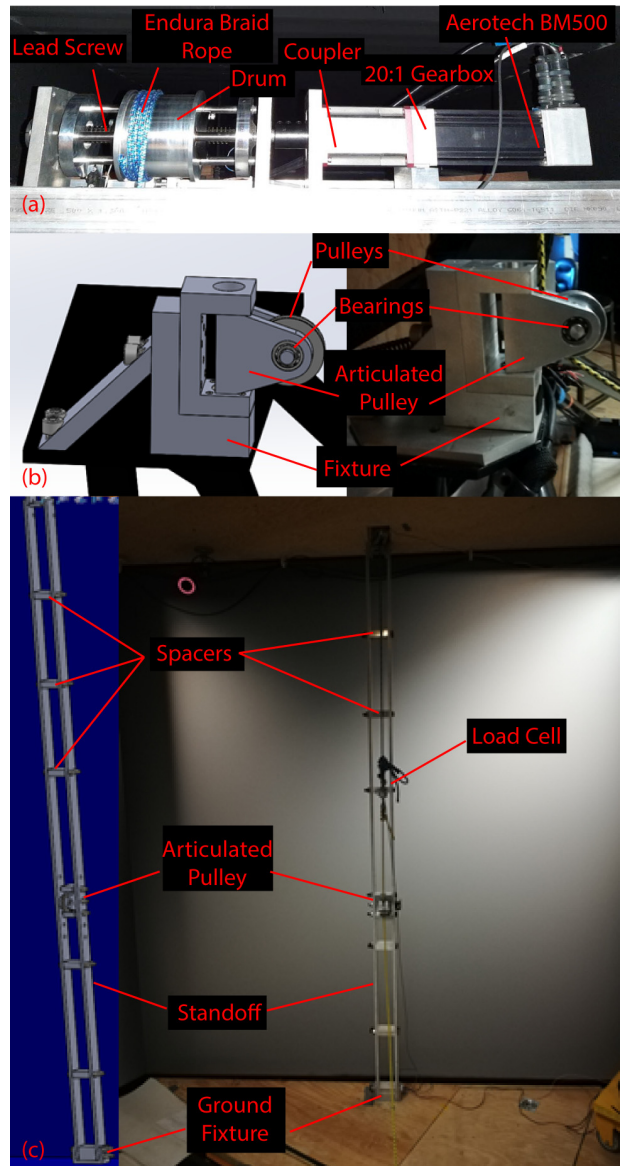

Fig. 2 (a) Motor and drum design. (b) CAD (left) and physical back articulated pulley (right). (c) CAD (left) and physical standoff (right).

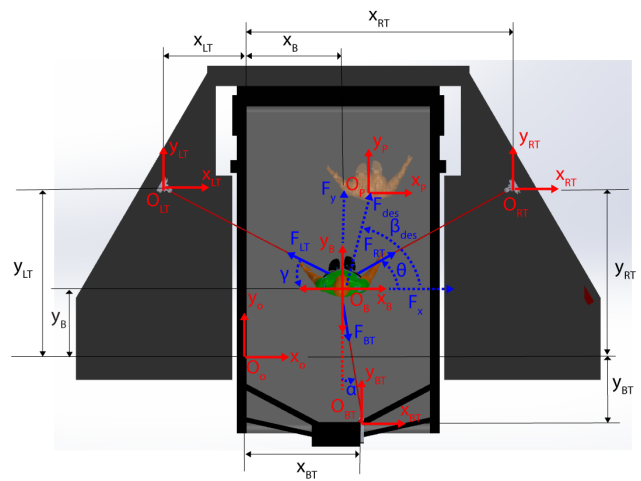

Fig. 3. System diagram indicating the position of the subject relative to the left, right, and back tether pulleys. The desired force and angle are shown with the subject being displaced

perturbation force and angle [7]. The resulting perturbation force is then,

$$F_x = F_{RT} \cos\theta - F_{LT} \cos\gamma + F_{BT} \sin\alpha$$
$$F_y = F_{RT} \sin\theta + F_{LT} \sin\gamma - F_{BT} \cos\alpha$$

where $F = \sqrt{F_x^2 + F_y^2}$ and $\beta = \text{atan}(F_y/F_x)$.

### C. Determining Nominal Tether Force

A nominal force on each of the tethers keeps the tether tight to minimize slack, prevents transverse vibration of the tether due to the runner's motion, and does not affect the mobility of the person. A 30 N nominal tension was determined empirically to minimize tether motion and not impede user motion. This tension is verified by examining the natural frequency of a vibrating string model,

$$\omega = \frac{n\pi c}{l} \quad c = \sqrt{\frac{\tau}{\rho A}}$$

where $n$ is the mode, $l$ is the tether length, $\tau$ is the tether tension, $\rho$ is the density, $A$ is the cross-sectional area, $m$ is the tether mass, and $\omega$ is the natural frequency in radians per second. Combining equations results in,

$$\omega = \frac{n\pi}{l}\sqrt{\frac{\tau l}{m}} \quad f = \frac{w}{2\pi} = \frac{n}{2l}\sqrt{\frac{\tau l}{m}}$$

where $f$ is the natural frequency in Hertz.

The goal is to maintain sufficient tension such that the first mode of vibration (i.e., n=1) is sufficiently above the frequency of excitation. Fig. 4 indicates how natural frequency changes with respect to tether tension. The 30 N nominal force correlates to a first mode (n=1) natural frequency of 8.75 Hz. Given that running is usually at a 3.57 Hz pace [19], the nominal tether tension is confirmed to be sufficient since it is ~2.5 times higher than the frequency of a person running.

### D. Workspace Analysis

It is important to apply a force to the user that is proportional to their mass so the perturbations are scaled consistently across users. Since participant mass varies, it is important to understand how the workspace of the system varies with force applied. To determine the workspace, a force sweep was conducted at different user positions. A desired force of 100, 200, 300, and 400 N was analyzed. Relative to a participant weighing 1000 N, this is equivalent to a 10%, 20%, 30%, and 40% scaled perturbation. The workspace analysis was used to determine the appropriate percentage of the person's body weight to apply safely on the participants given the dimensions of the treadmill.

Given the perturbation forces (i.e., 100, 200, 300, and 400 N), the workspace area of the system is 2.93, 1.86, 1.34, and 0.42 $m^2$ respectively. Fig. 5 illustrates the size and shape of the workspaces. The workspace was generated by discretizing the TPAWT area and checking that the system can provide omnidirectional forces at each discrete position. The blue dots show where the system can provide omnidirectional perturbations forces. The workspace is centered near the middle of the treadmill and becomes smaller with larger perturbation forces.

A perturbation force of 30% of the participant's body weight was selected for the perturbations. This correlates to the 300 N load in Fig. 5 for a 1000N participant weight. The workspace shows large resulting coverage of the treadmill area. The perturbation was verified to result in notable perturbations that participants could withstand safely.

### E. Motion Tracking and System Synchronization

It is important to apply perturbation forces at the same moment in each participant's locomotor cycle so that experimental results are consistent. In this case, perturbation forces are applied relative to when the participant's right foot is in the stance position. A Vicon Bonita system operating at 200 Hz tracks markers on the toe, heal and ankle, Fig. 6. A marker set on the user's waist harness tracks their position in the workspace, Fig. 1. Marker positions are then broadcasted by a virtual-reality peripheral network (VRPN) and processed by the tether control computer. Waist position data are used to determine appropriate tether forces given user position, Fig. 3. Foot position data (i.e., ankle, toe, and heal markers) are used to track locomotor cycle, predict when stance will occur and provide a priori warnings, and consistently apply perturbation forces in synchronization with the stance phase of the participant's stride.

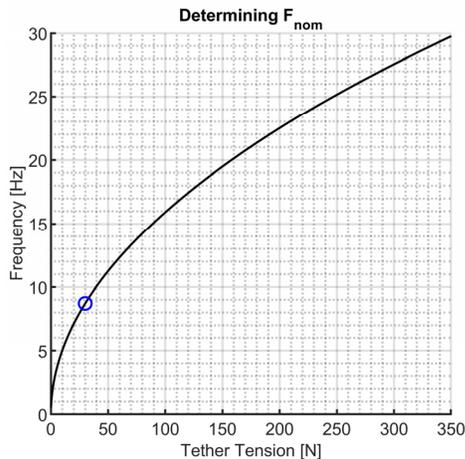

Fig. 4. Tether tension versus first natural frequency. The selected nominal 30 N force is shown by the blue circle.

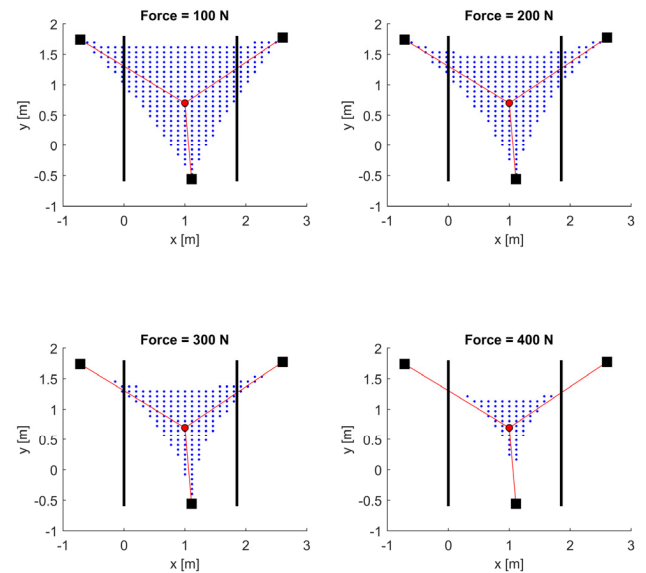

Fig. 5. Effective workspace overlaid on the treadmill with pulley, tether, and user position using 100, 200, 300, and 400 N on a 1000 N participant.

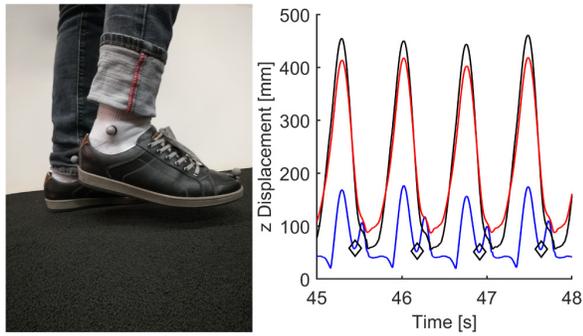

Fig. 6. Real-time gait detection based upon mid-stance indicated by diamond markers. The black signal is the ankle marker, the red signal is the heel marker, and the blue signal is the toe marker.

*F. Gait Algorithms*

The point in the gait we are interested in is when the participant's legs pass each other (i.e., stance). When right foot stance occurs, the participant's legs pass each other, and the right foot is planted so the participant can prepare for the perturbation.

An algorithm is implemented for detecting right foot stance instances during gait in real-time [7]. The **stance detection algorithm** tracks the z-position of the ankle, toe, and heel markers and pinpoints when stance occurs. The algorithm works by detecting when 1) the ankle marker and the heel marker pass each other with a negative slope, and 2) then finds the point where the toe height variation has a positive slope, Fig. 6.

*G. Predictive Warnings*

The warning system was designed to provide directional warnings by indicating the applied force direction before it is applied. The audio warning system consists of a series of piezoelectric buzzers located around the participant. Buzzers are located ~2 m to the left and right and in front and back of the participant but are offset to assist in sound localization,

Fig. 1. The rear buzzer emits a different tone to help discern front and back warnings. The back buzzer emits a frequency of 2 kHz for a period of 0.06 sec, where the right, front, and back buzzers emit a frequency of 1 kHz for a period of 0.03 sec.

The visual warning system is located directly in front of the participant at chest height about 2 m away, Fig. 1. The size of the indicator is 232.3 cm$^2$. It has LED light clusters indicating the four directions the participant will be perturbed from. These buzzers and LED lights turn on at the same instance the different warning type occurs. As mentioned before, warnings are either only auditory warnings, only visual warnings, or combined audio-visual warnings to help the participant discern the direction of the upcoming perturbation.

Warnings are triggered in advance of the applied force to give the participant reaction time to prepare for the perturbation. As such, the gait synchronization system maintains a running average of the user's gait, which is used to predict precisely when the perturbation will be applied and the time the alarm should be signaled. Based upon [20], the warnings are triggered 500 ms before the perturbation is applied, allowing sufficient time to respond, but not so much time as to provide a tactical advantage during a sport.

*H. Full System Logic*

The logic of the full autonomous system is shown in Fig. 7. This figure shows how the whole system is integrated together and the decision process the system makes to create a perturbation. Multiple operators are required to conduct the study. The first operator controls the treadmill computer and VICON computer. The treadmill computer regulates belt speed using a PID speed controller based upon belt motor encoder feedback. The VICON computer continuously tracks markers attached to the user to collect data and broadcast data over the ethernet via a VRPN. The VRPN sends the right foot toe, heel, and ankle marker positions and the user position cluster to the tether motor computer.

The second operator controls the tether motor computer. The tether computer first creates a randomized perturbation type with the perturbation type selector choosing between a warned or unwarned perturbation at toe-off or stance going either left/right/front/back. Only stance is examined here since it allows the user to best brace for simulated impacts. Once a randomized time (15-25 seconds) passes between perturbations, the warning will be activated 500 ms before the next perturbation using the running average of the locomotor cycle period received from the gait prediction algorithm. The gait algorithms use the right foot markers to pinpoint events. The warning will occur, then the perturbation will be activated once the participant hits the desired point in their gait. The perturb block is then activated to send the perturbation.

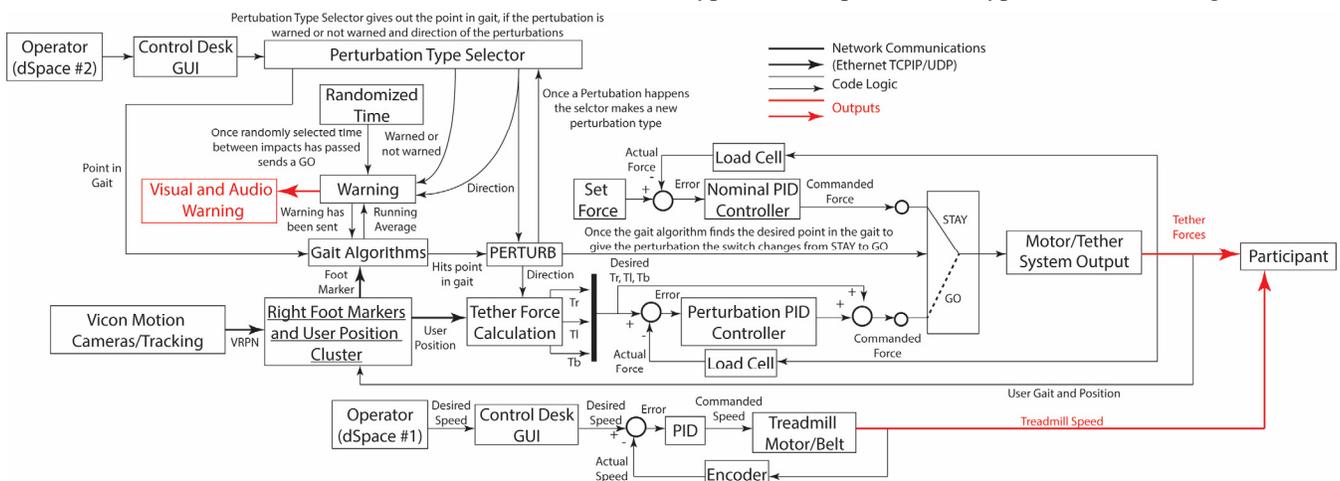

Fig. 7. Full System Logic

Each of the tether motors uses the same control scheme when a force is sent. The motor commands are calculated as torques, which directly relate to the current being sent to the motors by the current control amplifiers. Load cells in the tethers provide sensor feedback to calculate the error between the desired and actual force for the controller. A PID feedback controller is applied for both the nominal force and perturbation force. The perturbation controller also has a feed-forward term to command the motors to apply the desired tether force directly, resulting in a faster response. The PID controller then improves the accuracy. The gains of the PID controllers are designed differently depending on whether the force applied is the nominal or perturbation force.

The perturb block resets the perturbation type selector, tells the tether force calculator the direction of the perturbation to calculate the appropriate tether forces using the position of the user, and tells the system to switch between the nominal tether force to the perturbation forces. The forces are applied to the participant and perturb them with respect to the type of perturbation chosen by the selector. VICON tracks the participant, closing the loop.

There are physical emergency stop switches that halt the treadmill and the perturbation motor system. They are held by observers at all times during a test. If pressed, they will stop the whole system.

### III. SYSTEM VALIDATION

This section presents experimental results designed to validate that performance of the system is sufficient for the subject study. Due to motion of the participant, consistency of perturbation forcing is validated by comparing forces applied to a stationary object and then to a running participant. Modeling the system provides validation of the mechanical bandwidth. The gait detection algorithm is evaluated to assure that warnings and perturbation forces are delivered at the correct instant during the participant's locomotor cycle.

#### A. Force Response: Stationary Object vs Running Participant

Force control is first verified to provide similar results when applied to a stationary object (i.e., a cement anchor strapped to the treadmill frame) and when a person is running. Perturbation forces were applied to the stationary object and a runner in multiple directions (β = 0°, 90°, 180°, 270°) correlated to right, forward, left, and backward directions. Typical force profiles received from the load cells and the average resulting perturbation force of three resultants are shown in Fig. 8 (left) for the stationary object. The average percent overshoot for a stationary object and the time required for each motor to arrive at its commanded value are presented in Table 1. A 125 N perturbation force was commanded for $\delta t = 250$ ms, but due to the ~55 ms rise time, the peak force was applied for ~195 ms. Overshoot of ~4% occurs, which is a result of controller tuning to attain a fast rise time. Slight ripple (~6N) occurs during peak forces, but the resulting force error is only ~3 N. Note that a constant 30 N tether force of is commanded before and after the impulse period to maintain tether tension, which is quite steady due to the stationary object.

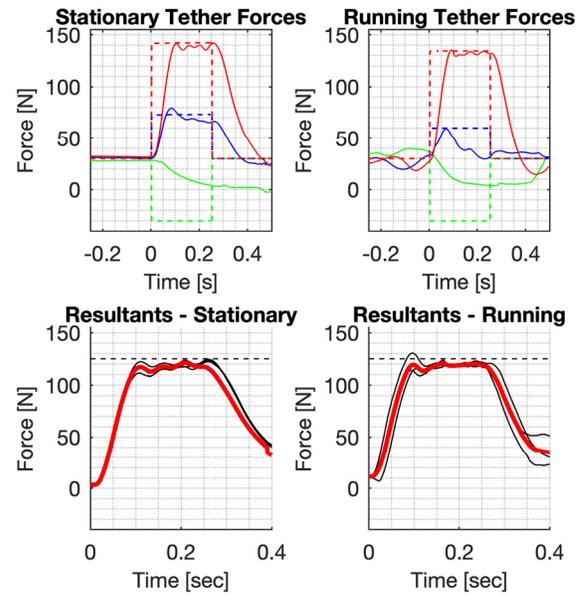

Fig. 8. Top: Tether forces applied to a stationary object (left) and a running person (right). Plots illustrate desired (dashed) versus actual (solid) tether force (blue=back, red=right, green=left). Bottom: Three resultant force profiles (black) with the average profile of those three profiles (red) for a stationary object (left) and a running person (right).

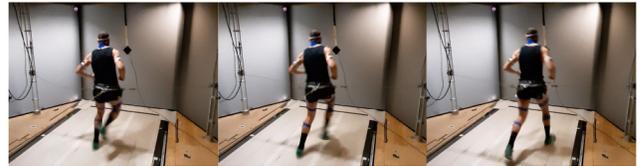

Fig. 9. Image sequence showing a person perturbed to the right while running

Tether force response and three averaged resultant force profiles, Fig. 8 (right), were also evaluated while a participant was running at 3 m/s. The perturbation force is commanded for the same duration as before ($\delta t$ = 250 ms). The motion of a user in response to the perturbation force is shown in Fig. 9, highlighting that resulting user motion is quite notable. The average percent overshoot for each motor and the time required to arrive at its commanded value are shown in Table 1. Rise time is longer (~100 ms) and overshoot is increased slightly (4.7%). Ripple is increased to ~12 N, but average peak force error is still small (~6N). Due to the rise time and user motion, the peak force is applied for ~150 ms. A constant tether tension of 30 N is again commanded before and after the perturbation, but forces vary due to user motion and rigidity of the stationary object, making coordination more difficult. Overall, forces are reasonable with variations attributed to user motion.

Table 1. Motor Responses for a Stationary Object and Running Participant with a right perturbation applied.

| Response | Stationary | | Running | |
|---|---|---|---|---|
| Motor | %OS | Rise Time [ms] | %OS | Rise Time [ms] |
| Right | 3.70% | 62 | 5.10% | 100 |
| Left | 4.20% | 60 | 4.60% | 110 |
| Back | 4.10% | 51 | 4.40% | 100 |

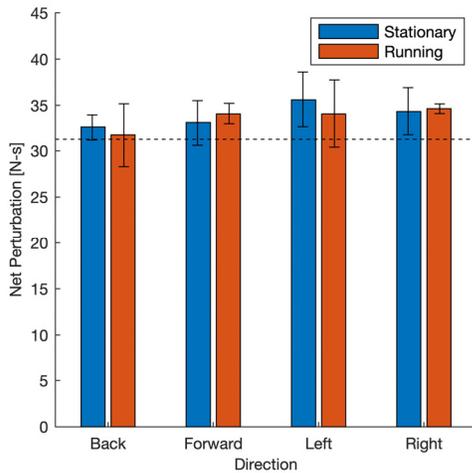

Fig. 10. Bar graph showing average perturbation impulses in each direction with error whiskers showing +/- two standard deviations. The desired impulse of 31.25 N-s is shown by the dashed line.

While there were variations in forces, the resulting applied impulses are close to the desired 31.25 N-s impulse. Fig. 10 shows the net impulse in each direction for a stationary object and runner. Due to the rise and fall of the peak forces, the net impulse was reasonably close to the desired values. On average, impulses were 2.3 N-s higher (7.5%) than desired for a running participant and 2.6 N-s (8.3%) higher than desired for the stationary object. Average standard deviation was also higher for the stationary object (2.4 N-s) compared to the running participant (2.1 N-s). The running participant introduced tether motion and allowed the person to move as forces were applied, whereas the stationary object was rigid and did not move. These results indicate that motion of the participant due to running did not have an adverse effect on performance of the system, and if anything, their compliance slightly reduced variation in the applied impulse.

Overall, performance of the forcing system was judged to be sufficient for proceeding with the subject study. Peak force errors were small and impulses matched sufficiently that repeated testing produced low average error.

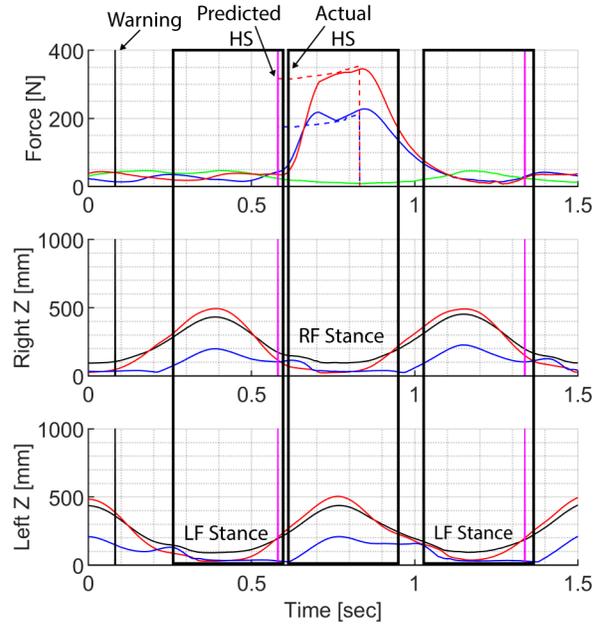

Fig. 11. Plots showing synchronization of forces, warnings, and right and left foot stance periods. Force plots (top) illustrate desired (dashed) versus actual (solid) tether forces (blue=back, red=right, green=left). The solid black vertical lines are when warnings are provided. The pink vertical line is where heal strike occurs. In the right foot and left foot plots, the black signal is the ankle marker, red is the heel marker, and blue is the toe marker.

### B. Force and Warning During Gait

Precise timing of warnings and force application relative to the users' gait is important to assure that users have consistent perturbations. Gait is tracked and stance is targeted. To evaluate consistency of the tracking algorithm, four different users' gait were analyzed to verify the algorithm worked for different users. A total of twenty warnings and force impulses were applied (i.e., five per user). For each of the participants, the algorithm pinpointed stance and could predict when to trigger the warning. The prediction of when to trigger the warning is obtained by taking the average time between stance locations and then subtracting the desired warning time. This gives the system the ability to predict when the next stance instance will occur in the participant's gait and give the warning at the appropriate time. Fig. 11 shows where the warning occurs and then where the force is applied with respect to stance. The desired warning time was 500 ms whereas the actual average warning time was 537.1 $ms$ with a standard deviation of 132.5 ms. Note that the predicted right foot heel strike (i.e., pink vertical lines) occurs slightly before (~40 ms) stance, which allows the forces to ramp up in time to be applied during the stance phase.

### C. Mechanical Bandwidth

The mechanical bandwidth of the system was analyzed to assure that the forcing system could respond quickly enough. To do this, a mathematical analysis of the system was conducted to find the mechanical transfer function. Fig. 12 shows a diagram of the components used to derive the system's mechanical transfer function. The transfer function

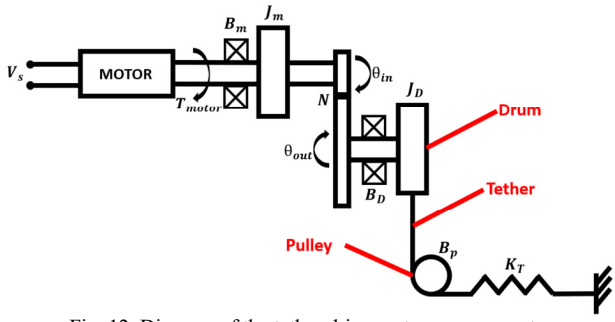

Fig. 12. Diagram of the tether drive system components.

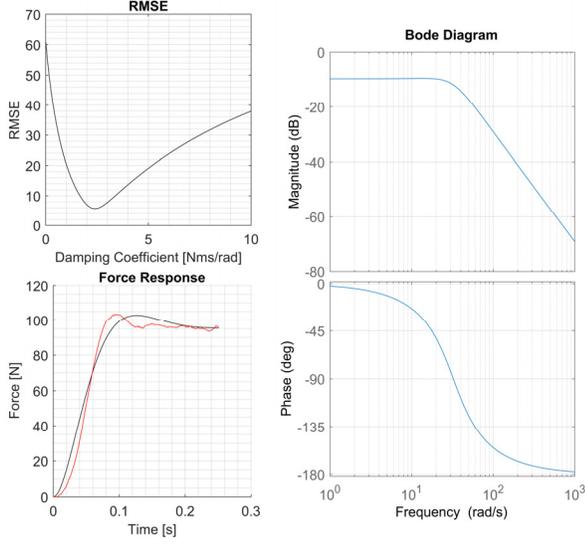

Fig. 13. Shows the RMSE response (top-left), force response (bottom-left): actual response = red, transfer function step response = black, and Bode diagram (right).

of the system is as follows:

$$\frac{\theta_{out}(s)}{T_m(s)} = \frac{N}{(N^2 J_m + J_D)s^2 + (N^2 B_m + B_p r_D + B_D)s + K_T r_D^2}$$

Where $N$ is the gear ratio, $J_m$ and $J_D$ are the motor and drum inertia respectively; $B_m$, $B_p$, and $B_D$ are the motor, pulley, and drum damping; $K_T$ is the tether spring constant, $r_D$ is the drum radius, $\theta_{out}$ is the output drum position, and $T_{motor}$ is the input motor torque.

The unknown variables are the damping terms $B_m$, $B_p$, and $B_D$. A 125 Newton perturbation was applied to a stationary object at the center of the treadmill by the back motor to determine those terms. The actual force profile and the system's step response were compared using Root Mean Squared Error (RMSE) and minimized to find the overall damping term. Fig. 13 shows the actual response and the transfer function step response. Bode diagrams are then used to visualize the magnitude and phase response of the system. The mechanical bandwidth of the system at -3 dB is 35.343 $\frac{rad}{s}$ or 5.625 Hz. This correlates to a 114 ms settling time as seen in Fig. 13. This value is also sufficient when compared to the bandwidth for tracking user motion during running as discussed before.

## IV. SUBJECT STUDY

This study demonstrates the ability of the system to provide different types of warnings to better understand how they help humans prepare for perturbations while running. These warnings are either an audio warning, visual warning, combined audio-visual warning or no warning at all prior to perturbation while running. Sub-sections describe the participant population and study design, setup, procedure, results, and discussion.

### A. Participants

Participants chosen for the full study were healthy adults capable of running 3 m/s for up to 15 minutes. Four males and two females participated in the study ranging in mass from 60 kg to 115 kg, and between the ages of 20 to 30. Participants are all highly active and athletic. Participants consented to this study under University of Utah IRB #132886.

### B. Design

The impact system was designed to perform the test autonomously. Maximum participant displacement after impact was used to assess the efficacy of the warnings. Impulses were applied during right foot stance, with $\beta = 0°$, $90°$, $180°$, and $270°$, either giving an audio, visual, combined audio-visual or no warning perturbation. All participants were subjected to 12 impacts for each of the different warning scenarios resulting in a total of 48 perturbation over 4-5 minutes, which allowed recovering time between perturbations. Each of the 4 different warning scenarios are tested separately, meaning the participant will be subjected to 12 perturbations of the specific warning type then the warning type will be changed. The order of warning types were randomized between participants. All impacts were given in a randomized sequence at time intervals ranging between 15-25 seconds. Applied force, $F_{des}$, was defined as 30% of the participant's body weight and commanded for $\delta t_{impulse} = 250$ ms at a constant velocity of 3 m/s.

### C. Procedures

Participants were equipped with a climbing harness and motion capture markers on the waist, and body tracking markers were placed on the toe, ankle, and heel. The participant was then led onto the treadmill belt where a researcher attached the system tethers to the harness. A practice session was conducted once the participant was in the system. Participants first experienced audio warning with respect to all directional warnings until they accurately identified the direction of each warning. The same was done for the visual and combined audio-visual warnings. The participant then experienced a set of warned impacts in four directions while standing to confirm whether the applied forces were acceptable. Unwarned perturbations were then applied to the participant to confirm that they were comfortable experiencing perturbations without warning.

Table 2. Linear Mixed Effect Model Results.

| LME Results | Right | | Front | | Left | | Back | |
|---|---|---|---|---|---|---|---|---|
| | Estimate [mm] | p-Value | Estimate [mm] | p-Value | Estimate [mm] | p-Value | Estimate [mm] | p-Value |
| Intercept (NW) | 314.1 | - | 387.7 | - | 379.1 | - | 398.5 | - |
| Audio Warning (A) | -28.3 | 0.008* | -36.8 | 0.120 | -89.0 | <0.001* | -36.1 | 0.129 |
| Visual Warning (V) | -65.9 | <0.001* | -50.1 | 0.035* | -136.4 | <0.001* | -128.2 | <0.001* |
| Audio Visual Warning (AV) | -56.7 | <0.001* | -37.3 | 0.114 | -137.0 | <0.001* | -114.5 | <0.001* |

After these training protocols were complete, participants were asked to resist perturbations while running at 3 m/s for all 4 of the different warned and unwarned scenarios. Once they were comfortable with running while experiencing perturbations, the experiment was conducted.

*D. Results and Discussion*

Using the body marker cluster described earlier, each perturbation direction displacement was analyzed. The gait detection system consistently determined stance, resulting in desired application of perturbations. For this study only stance was analyzed to give the participant the ability to react to the warning when having one leg planted. Both toe-off and mid-stance were analyzed in a pilot study [7], but are not reiterated here.

Each of the different directions were split up to examine the effects of the different warning scenarios all applied at stance. A Linear Mixed Effects Model was applied to the displacement data (dependent variable) with unwarned perturbation displacements being the null case. The warning type is the predictor variable (independent variable), and the participant is the random effects variable (Table 2). The Intercept estimate is the displacement with no warning (NW), whereas the estimates for audio warning (A), visual warning (V), and combined audio-visual (AV) warnings are the change in displacement magnitude relative to NW.

Using a significance level of $p < 0.05$ we see significant displacement reduction in nine of the twelve perturbation scenarios. Visual warnings showed significance in all directions, suggesting that participants only needed a visual cue to brace for the perturbation. Audio warnings showed significance for the right and left perturbations but not for the forward and back. A possible explanation is that humans have difficulty discerning audio direction of an acoustic signal when the source is located in the sagittal plane. Combined audio-visual warnings showed significance in all directions except for forward. This may suggest that the user was attentive to both audio and visual warnings, but audio may have added some ambiguity to the direction.

Fig. 14 shows box and whisker plots of the displacements separated into each of the different perturbation directions at stance. The plots show reduced displacement when warned for all warning types compared to the not-warned perturbation for all directions except for the front perturbation. Visual warnings have the lowest amount of displacement followed by combined audio-visual warnings, and lastly audio warnings.

Thus, the warning systems allowed participants to better prepare for a perturbation. By supplying an audio warning before the perturbation, the subject can prepare for a right or left perturbation. Supplying a combined audio-visual warning before a perturbation, a participant can prepare for a right, left, and back perturbation. Lastly a participant supplied with only a visual warning can prepare for a perturbation in all of the directions.

V. CONCLUSIONS

This article examines the efficacy of audio and visual warnings to help a user prepare for an impending perturbation force while they are running. Adaptation of the TPAWT is validated to create an effective three tether parallel robot for providing perturbation forces during the stance phase of stride. This is made possible by predictive gait tracking, which triggers warnings moments before perturbation forces are applied. Force response of the system is examined to assure that force application is sufficient during running. Workspace of the system is examined to assure that the resulting workspace is sufficient for perturbations scaled to user mass. A user study examines the effect of warnings (visual, audio, visual + audio, and no warnings) to help users prepare for perturbation forces from different directions. Visual warnings helped the user significantly reduce their displacements from all directions, whereas audio+visual warnings were significant in left, right, and back perturbations. Audio was only significant in left/right perturbation warnings. Reduction of displacements follows similar trends, with visual having the most effect and audio having the least. These results are important for researchers designing human-robot systems where communication of urgent events is critical, such as a Smart Helmet warning a user of an impending impact.

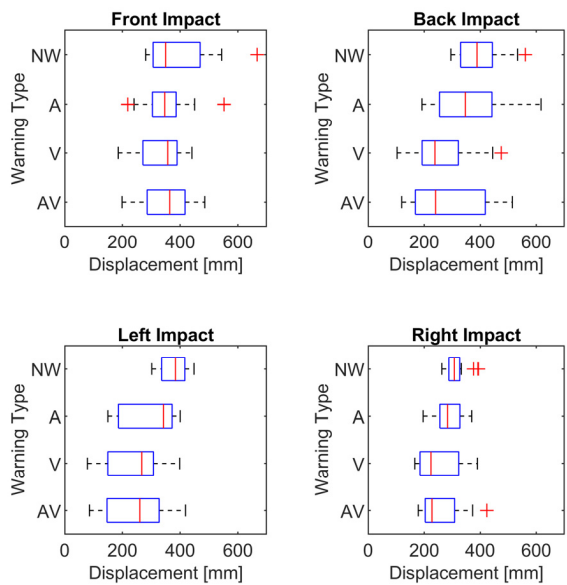

Fig. 14. Box and whisker plot showing displacements of the perturbed participants while running in the respected direction during stance for each warning scenario (NW- No Warning, A – Audio, V- Visual, AV – Audio + Visual).

**Nathaniel G. Luttmer** Department of Mechanical Engineering, University of Utah at Salt Lake City, Utah, 84112, USA. Email: nluttmer1@gmail.com.

**Takara E. Truong** Computer Science Department, Stanford University at Stanford, California, 94305, USA. Email: takaraet@gmail.com.

**Alicia M. Boynton** School of Biological Sciences at the University of Utah in Salt Lake City, Utah, 84112, USA. Email: Alicia.boynton@utah.edu.

**Andrew S. Merryweather** Department of Mechanical Engineering, University of Utah at Salt Lake City, Utah, 84112, USA. Email: a.merryweather@utah.edu.

**David R. Carrier** School of Biological Sciences at the University of Utah in Salt Lake City, Utah, 84112, USA. Email: carrier@biology.utah.edu.

**Mark A. Minor** Department of Mechanical Engineering, University of Utah at Salt Lake City, Utah, 84112, USA. Email: mark.minor@utah.edu.